\documentclass{article}
\usepackage{iclr2023_workshop,times}

\usepackage{amsmath,amsfonts,bm}

\def\eqref#1{equation~\ref{#1}}

\def\1{\bm{1}}

\DeclareMathAlphabet{\mathsfit}{\encodingdefault}{\sfdefault}{m}{sl}
\SetMathAlphabet{\mathsfit}{bold}{\encodingdefault}{\sfdefault}{bx}{n}

\usepackage{hyperref}
\usepackage{url}
\usepackage{array}
\usepackage{setspace}
\usepackage{diagbox}
\usepackage{mathrsfs}
\usepackage{makecell}
\usepackage{diagbox}
\usepackage{caption}
\usepackage{float}
\usepackage{amsmath}
\usepackage{amsthm}
\usepackage{amssymb}
\usepackage{tikz}
\tikzset{
  every neuron/.style={
    circle,
    draw,
    minimum size=2pt,
    inner sep=1pt,
    fill=black
  }
}

\title{Physics-inspired interpretability of machine learning models}

\author{Maximilian P.~Niroomand, David J.~Wales \\
Department of Chemistry \\
University of Cambridge\\
\texttt{\{mpn26,djw34\}@cam.ac.uk}}

\iclrfinalcopy 

\begin{document}

\maketitle

\begin{abstract}
The ability to explain decisions made by machine learning models remains one of the most significant hurdles towards widespread adoption of AI in highly sensitive areas such as medicine, cybersecurity or autonomous driving. Great interest exists in understanding which features of the input data prompt model decision making. In this contribution, we propose a novel approach to identify relevant features of the input data, inspired by methods from the energy landscapes field, developed in the physical sciences. By identifying conserved weights within groups of minima of the loss landscapes, we can identify the drivers of model decision making. Analogues to this idea exist in the molecular sciences, where coordinate invariants or order parameters are employed to identify critical features of a molecule. However, no such approach exists for machine learning loss landscapes. We will demonstrate the applicability of energy landscape methods to machine learning models and give examples, both synthetic and from the real world, for how these methods can help to make models more interpretable.   
\end{abstract}

\section{Introduction}
Machine learning methods have achieved impressive results in recent years. Besides famous applications in areas like chess \citep{silver2017mastering} and Go \citep{silver2017masteringgo}, AI plays a critical role in advances to autonomous driving \citep{grigorescu2020survey}, protein structure prediction \citep{jumper2021highly}, cancer identification \citep{sammut2022multi} and in cybersecurity \citep{dasgupta2022machine}. However, in order for AI methods to take the next step and be commonly employed for critical applications without any humans in the loop, we want to be able to understand the decision making process. A critical component towards explainable AI is understanding which parts of the input data are utilised by the model in its decision making. In neural networks, the most popular approach is to study the outgoing weights and gradients from an individual input node. Larger weights are reasonably assumed to indicate a greater significance of the particular input, and indeed, an entire class of interpretability metrics, namely gradient-based methods, are founded on this idea \citep{simonyan2013deep, linardatos2020explainable}. Yet, given the immense complexity of overparameterised, deep neural networks, current methods are in practice often insufficient to appropriately explain a model. Using methods from the physical sciences, we propose a novel approach as a next step towards interpretable neural networks. 

\subsection{Energy landscapes}
In the physical sciences, energy landscapes (ELs) are employed to explore molecular configuration space \citep{wales1998archetypal, wales2003energy}. Each molecular configuration is associated with an energy value, and local minima of the energy landscape represent stable isomers. The analogy to machine learning loss landscapes (ML-LLs) is straightforward, the main difference perhaps being that non-minima are valid configurations for sets of weights. Due to this similarity between ELs and ML-LLs, various, well-established methods from the field of energy landscapes can be employed to study ML-LLs. One key area of interest here is interpretability. Employing well-understood methods from a mature field, with a solid mathematical basis in the physical world, to move away from black-box machine learning models may be a helpful step towards interpretable machine learning models.

\subsection{Related work}
Various approaches to interpretability in deep learning for neural networks exist. Below, we are mostly interested in gradient-based methods due to their applicability to non-image data. Various other methods to interpret the output of CNNs on images exist, as for example summarised in \citet{linardatos2020explainable}, but will not be reviewed below.  
\newline \textbf{Gradient-based methods:} 
All gradient-based methods are concerned with changes in the prediction as the input data is slightly perturbed. For a vector-valued input $\mathbf{x} \subset \mathbf{\mathrm{X}} \in \mathbb{R}^d$ and some loss function, $\mathcal{L}$, a gradient-based method computes some expression of the form $\partial \mathcal{L}/ \partial \mathbf{x}$, usually for each input node individually. 
Gradient-based methods were first introduced for images by \citet{simonyan2013deep}, who used them to compute how changes in the input affect predictions in the neighbourhood of the input, allowing the computation of a salience map \citep{kummerer2014deep, zhao2015saliency}. More recently, integrated gradient methods \citep{sundararajan2017axiomatic} consider the derivative of the output (loss) with respect to individual input nodes. If the change in loss is large with respect to some input feature, that feature is more likely to be relevant to the decision making. 
Various other gradient and perturbation based methods exist \citep{alvarez2018robustness}, yet their usefulness and accuracy is debated, and is generally agreed to be insufficient \citep{srinivas2020rethinking}. 
\newline\textbf{Energy landscapes in machine learning:}
Energy landscapes methods have been employed to study machine learning in previous contributions \citep{ballard2017energy, chitturi2020perspective}. \citet{niroomand2022capacity} used energy landscapes to characterise new loss functions, and the landscapes view has been used more broadly to gain insights into machine learning models \citep{segura2022subaging, verpoort2020archetypal}. Lastly, other applications of energy landscape methods have employed various concepts from physical sciences in machine learning, including the heat capacity \citep{bradley2022shift, niroomand2022capacity}, both for characterisation and model improvement.
\newline\textbf{Interpreting energy landscapes:}
Due to the associated physical meaning, energy landscapes are usually more easily interpretable. Only minima represent equilibrium configurations, and each minimum is associated with a unique structure. However, for larger, complex molecules, many minima may exist, and enumerating them may be infeasible. Instead, common features between sets of minima, grouped by their energetic properties, may be identified. For example, in \citep{roder2020structural} and \citep{roder2022energy} a multi-funnelled landscape is analysed to understand which structural differences of a molecule characterise  solutions in a specific funnel.

\section{Energy landscapes methods}
The study of energy landscapes is a well-established field \citep{wales1998archetypal}. Various approaches exist for constructing a faithful representation of the landscape by optimising the non-convex energy function, and visualising this landscape. Visualisation is commonly performed using disconnectivity graphs \citep{becker1997topology, wales1998archetypal} as described below. 
\subsection{Landscape visualisation}
A disconnectivity graph is a low-dimensional representation of a complex function landscape, which reduces the function to key characteristic stationary points, namely minima and transition states. A transition state is an index-1 saddle point of the funciton. The vertical axis of a disconnectivity graph represents the energy or loss value, and ordering along the horizontal axis is arbitrary. To identify distinct groups of minima, usually called funnels, we introduce the notion of levels and nodes. Levels are cross-sections of the energy at some evenly spaced, discrete heights in the disconnectivity graph. The highest energy level in the disconnectivity graph is level 1, and the lowest corresponds to the global minimum. Thus, each minimum belongs to one of evenly spaced intervals. Within each level, minima are grouped by a shared parent node, located higher up. In the disconnectivity graphs below, levels and nodes are represented as level\_node.

In the molecular sciences, a transition state between two minima describes the energy barrier to be overcome for a molecule to change configuration from one state to the other. This particular notion does not have a direct meaning in machine learning. However, given the optimisation procedure required for model training, the concept of a transition state is highly relevant, since it may determine which minimum basin the optimiser will fall into. Thus, disconnectivity graphs can be employed as a faithful coarse-grained representation of the loss landscape. In particular, it will be relevant below to understand that any group of minima close together, perhaps separated from other groups of minima via high-lying transition states, may share commonalities. This effect has been observed in \citep{roder2020structural, roder2022energy} for molecular systems, and we find that the same argument holds for ML-LLs.

\section{Experiments}
\begin{figure}[!h]
    \centering
\begin{tikzpicture}
    \node[anchor=south west,inner sep=0] (image) at (0,0) {\includegraphics[width=0.7\textwidth]{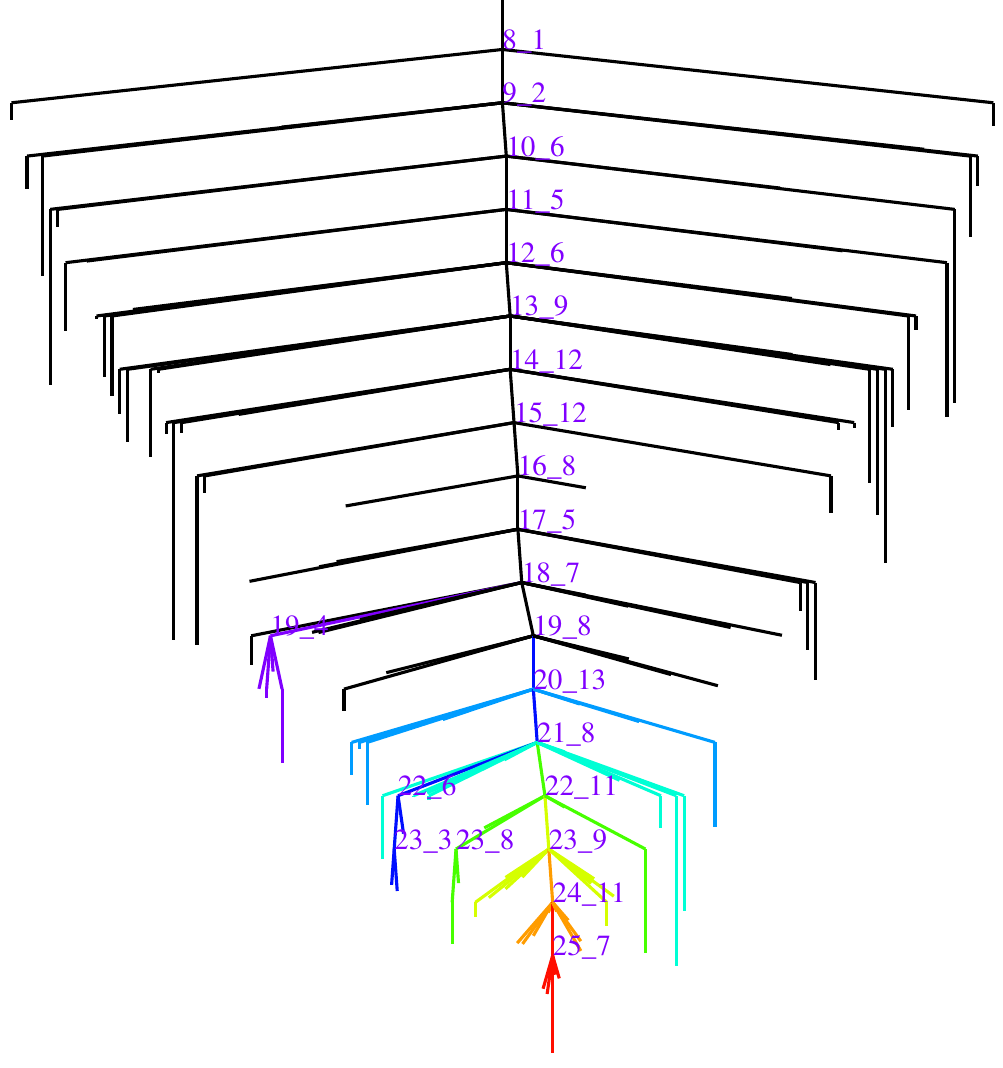}};
    \begin{scope}[xshift=5.7cm, yshift=-0.6cm]
        \foreach \m [count=\y] in {1,2}
          \node [every neuron/.try, neuron \m/.try] (input-\m) at (0,1-\y*0.2) {};
        \foreach \m [count=\y] in {1,2,3,4,5}
          \node [every neuron/.try, neuron \m/.try ] (hidden-\m) at (0.5,1.3-\y*0.2) {};
        \foreach \m [count=\y] in {1,2}
        \node [every neuron/.try, neuron \m/.try ] (output-\m) at (1,1-\y*0.2) {};
        \foreach \i in {1,2}
          \foreach \j in {1,2,3,4,5}
            \draw [-] (input-\i) -- (hidden-\j);   
        \foreach \l [count=\i] in {1,2,3,4,5}
          \foreach \k [count=\j] in {1,2}
            \draw [-] (hidden-\i) -- (output-\j);
        \draw [-,red, ultra thick] (input-2) -- (hidden-2);
        \draw [-,red, ultra thick] (input-1) -- (hidden-4);
        \draw [-,red, ultra thick] (input-2) -- (hidden-4);
        \draw [-,red, ultra thick] (hidden-1) -- (output-1);
        \draw [-,red, ultra thick] (hidden-2) -- (output-2);
        \draw [-,red, ultra thick] (hidden-3) -- (output-2);
        \draw [-,red, ultra thick] (hidden-5) -- (output-1);
    \end{scope}

    \begin{scope}[xshift=2.9cm, yshift=0.6cm]
        \foreach \m [count=\y] in {1,2}
          \node [every neuron/.try, neuron \m/.try] (input-\m) at (0,1-\y*0.2) {};
        \foreach \m [count=\y] in {1,2,3,4,5}
          \node [every neuron/.try, neuron \m/.try ] (hidden-\m) at (0.5,1.3-\y*0.2) {};
        \foreach \m [count=\y] in {1,2}
        \node [every neuron/.try, neuron \m/.try ] (output-\m) at (1,1-\y*0.2) {};
        \foreach \i in {1,2}
          \foreach \j in {1,2,3,4,5}
            \draw [-] (input-\i) -- (hidden-\j);   
        \foreach \l [count=\i] in {1,2,3,4,5}
          \foreach \k [count=\j] in {1,2}
            \draw [-] (hidden-\i) -- (output-\j);
        \draw [-,blue, ultra thick] (input-1) -- (hidden-3);
        \draw [-,blue, ultra thick] (input-2) -- (hidden-3);
        \draw [-,blue, ultra thick] (input-2) -- (hidden-4);
        \draw [-,blue, ultra thick] (input-2) -- (hidden-5);
        \draw [-,blue, ultra thick] (hidden-2) -- (output-1);
        \draw [-,blue, ultra thick] (hidden-2) -- (output-2);
        \draw [-,blue, ultra thick] (hidden-4) -- (output-2);
        \draw [-,blue, ultra thick] (hidden-5) -- (output-1);
    \end{scope}

    \begin{scope}[xshift=1.8cm, yshift=1.8cm]
        \foreach \m [count=\y] in {1,2}
          \node [every neuron/.try, neuron \m/.try] (input-\m) at (0,1-\y*0.2) {};
        \foreach \m [count=\y] in {1,2,3,4,5}
          \node [every neuron/.try, neuron \m/.try ] (hidden-\m) at (0.5,1.3-\y*0.2) {};
        \foreach \m [count=\y] in {1,2}
        \node [every neuron/.try, neuron \m/.try ] (output-\m) at (1,1-\y*0.2) {};
        \foreach \i in {1,2}
          \foreach \j in {1,2,3,4,5}
            \draw [-] (input-\i) -- (hidden-\j);   
        \foreach \l [count=\i] in {1,2,3,4,5}
          \foreach \k [count=\j] in {1,2}
            \draw [-] (hidden-\i) -- (output-\j);
        \draw [-,violet, ultra thick] (input-2) -- (hidden-2);
    \end{scope}

\end{tikzpicture}
    \caption{Disconnectivity graph for the checkerboard dataset. The conserved weights for a specific local minimum are highlighted in the respective colour for the chosen examples.}
    \label{fig:cb_dg}
\end{figure}
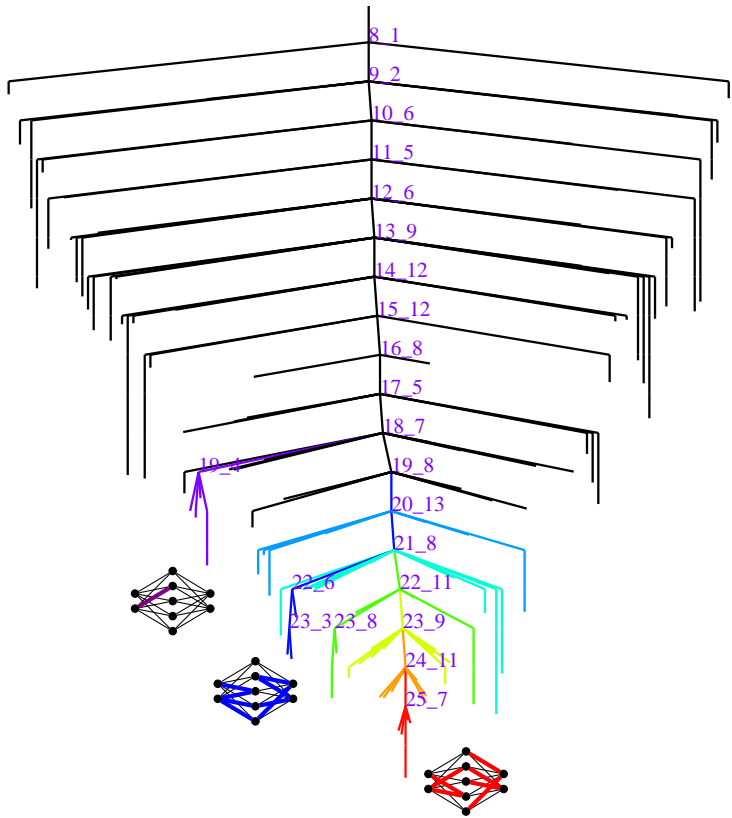

\begin{figure}[!h]
    \centering
\begin{tikzpicture}
    \node[anchor=south west,inner sep=0] (image) at (0,0) {\includegraphics[width=0.6\textwidth]{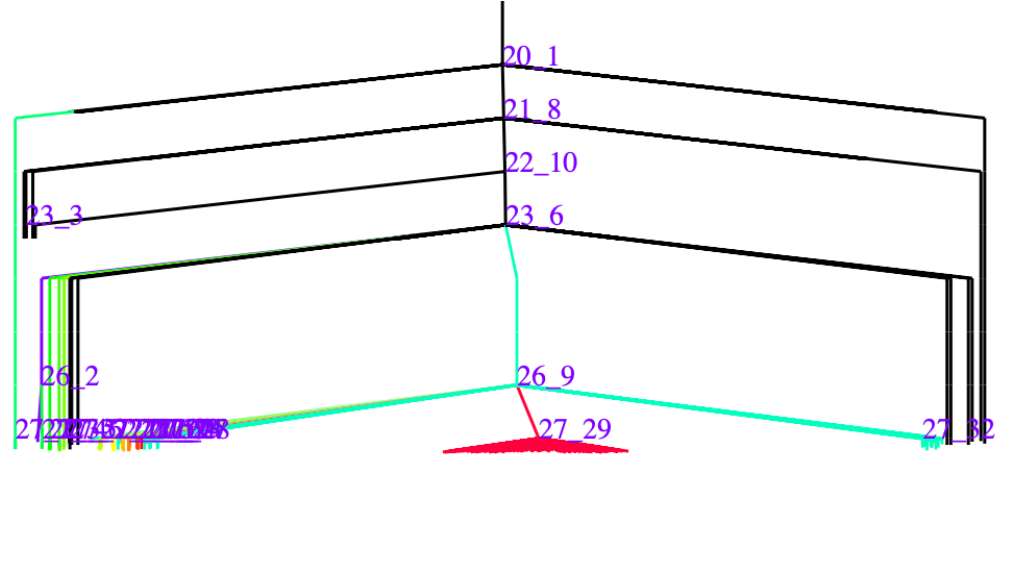}};
    \begin{scope}[xshift=10cm, yshift=3cm]
        \foreach \m [count=\y] in {1,2,3,4,5,6,7,8,9,10,11,12,13,14,15,16,17,18,19,20,21,22,23,24,25,26,27,28,29}
          \node [every neuron/.try, neuron \m/.try] (input-\m) at (0,4.15-\y*0.2) {};
        \foreach \m [count=\y] in {1,2,3}
          \node [every neuron/.try, neuron \m/.try ] (hidden-\m) at (1,2-\y*0.4) {};
        \foreach \m [count=\y] in {1,2}
        \node [every neuron/.try, neuron \m/.try ] (output-\m) at (2,1.8-\y*0.4) {};
        \foreach \i in {1,2,3,4,5,6,7,8,9,10,11,12,13,14,15,16,17,18,19,20,21,22,23,24,25,26,27,28,29}
          \foreach \j in {1,2,3}
            \draw [-] (input-\i) -- (hidden-\j);   
        \foreach \l [count=\i] in {1,2,3}
          \foreach \k [count=\j] in {1,2}
            \draw [-] (hidden-\i) -- (output-\j);
        \draw [-,red, ultra thick] (input-6) -- (hidden-1);
        \draw [-,red, ultra thick] (input-6) -- (hidden-2);
        \draw [-,red, ultra thick] (input-6) -- (hidden-3);
    \end{scope}
    \draw [red] (4.8,1.3) -- (9.7, 4.3);

    \begin{scope}[xshift=3cm, yshift=5cm]
        \foreach \m [count=\y] in {1,2,3,4,5,6,7,8,9,10,11,12,13,14,15,16,17,18,19,20,21,22,23,24,25,26,27,28,29}
          \node [every neuron/.try, neuron \m/.try] (input-\m) at (4.15-\y*0.2,2) {};
        \foreach \m [count=\y] in {1,2,3}
          \node [every neuron/.try, neuron \m/.try ] (hidden-\m) at (2-\y*0.4,1) {};
        \foreach \m [count=\y] in {1,2}
        \node [every neuron/.try, neuron \m/.try ] (output-\m) at (1.8-\y*0.4,0) {};
        \foreach \i in {1,2,3,4,5,6,7,8,9,10,11,12,13,14,15,16,17,18,19,20,21,22,23,24,25,26,27,28,29}
          \foreach \j in {1,2,3}
            \draw [-] (input-\i) -- (hidden-\j);   
        \foreach \l [count=\i] in {1,2,3}
          \foreach \k [count=\j] in {1,2}
            \draw [-] (hidden-\i) -- (output-\j);
        \draw [-,green, ultra thick] (input-6) -- (hidden-1);
        \draw [-,green, ultra thick] (input-11) -- (hidden-1);
        \draw [-,green, ultra thick] (input-13) -- (hidden-1);
        \draw [-,green, ultra thick] (input-14) -- (hidden-1);
        \draw [-,green, ultra thick] (input-18) -- (hidden-1);
        \draw [-,green, ultra thick] (input-24) -- (hidden-1);
        \draw [-,green, ultra thick] (input-6) -- (hidden-2);
        \draw [-,green, ultra thick] (input-7) -- (hidden-2);
        \draw [-,green, ultra thick] (input-13) -- (hidden-2);
        \draw [-,green, ultra thick] (input-17) -- (hidden-2);
        \draw [-,green, ultra thick] (input-18) -- (hidden-2);
        \draw [-,green, ultra thick] (input-24) -- (hidden-2);
        \draw [-,green, ultra thick] (input-2) -- (hidden-3);
        \draw [-,green, ultra thick] (input-3) -- (hidden-3);
        \draw [-,green, ultra thick] (input-6) -- (hidden-3);
        \draw [-,green, ultra thick] (input-7) -- (hidden-3);
        \draw [-,green, ultra thick] (input-9) -- (hidden-3);
        \draw [-,green, ultra thick] (input-11) -- (hidden-3);
        \draw [-,green, ultra thick] (input-17) -- (hidden-3);
        \draw [-,green, ultra thick] (input-18) -- (hidden-3);
        \draw [-,green, ultra thick] (input-21) -- (hidden-3);
        \draw [-,green, ultra thick] (input-21) -- (hidden-3);
        \draw [-,green, ultra thick] (input-24) -- (hidden-3);
        \draw [-,green, ultra thick] (input-28) -- (hidden-3);
    \end{scope}
    \draw [green, thick] (0.75,2.6) -- (4, 4.8);

\end{tikzpicture}
\caption{Disconnectivity graph for credit card data. Group 25\_7 in red includes the global minimum. Coloured edges indicate that for all minima in the specific group, these particular weights are conserved, i.e. have a standard deviation $< n$ which has been set to $n = 0.01$ here.}
\label{fig:cc_dg}
\end{figure}
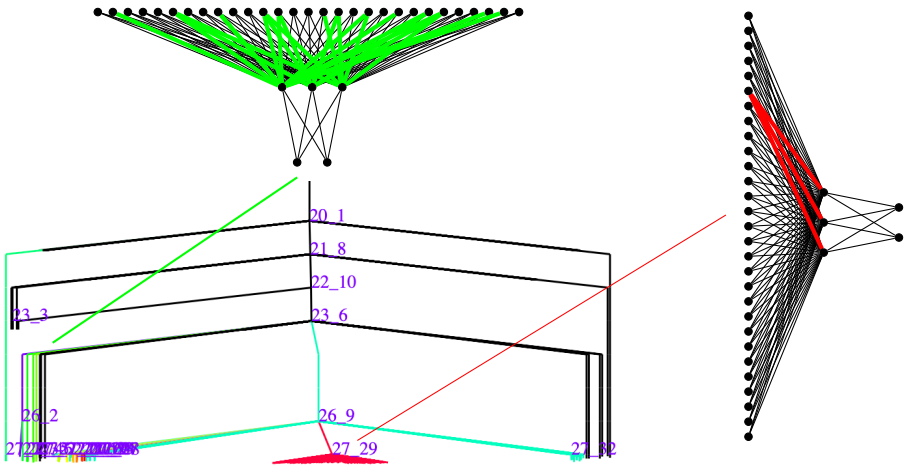

We report results for two separate experiments on two datasets. We believe that the underlying idea applies without loss of generality to any neural network architecture. However, further work will be required to validate this suggestion. Figures \ref{fig:cb_dg}and \ref{fig:cc_dg} show disconnectivity graphs for (1) a 2-dimensional synthetic checkerboard dataset \citep{kluger2003spectral} and (2) an anonymised, 29-dimensional credit card fraud detection dataset \citep{dal2015calibrating}, which are binary classification problems. The lowest lying node in each graph is the global minimum. To identify groups of minima with conserved weights, we follow a two-step procedure. Firstly, we identify groups of minima, that are separated from other groups by a higher-lying transition state. This segregation leads to the notion of nodes and levels described above. Secondly, we identify groups of minima that share a subset of conserved weights by computing the standard deviation of each weight across each node in each level. A subset of weights $\tilde w \subset W$ is conserved if $\sigma(w) < n$ for any $w \in W$, where $W$ denotes the weights of all minima in one node of one level. 

For visualisation purposes, we employ single-layer neural networks, which is sufficient for our analysis, with only a few nodes. The AUC of the best solutions is $> 0.95$ for both problems. Hence, these networks provide a realistic solutions to the set problems. In both figures, we visualised the conserved weights for a group of minima in the corresponding colour. In Figure \ref{fig:cb_dg}, various weights across the network are conserved, highlighting how this approach identifies relevant weights for the model. 
In Figure \ref{fig:cc_dg}, the funnel containing the global minimum (red) conserves 3 weights, all related to one specific input node. Randomly permuting the 3 identified weights for the group of minima around the global minimum in figure \ref{fig:cc_dg} reduces the best AUC from $\approx0.95$ to 0.76. In contrast, permuting any random set of 3 weights by the same magnitude on average only decreases the best AUC by 0.05 to an average best AUC of $\approx0.9$. In \ref{fig:cc_dg}, for group 25\_7 (red), weights for only a single minimum are conserved, in group 22\_6 (blue), weights outgoing from different input nodes are conserved. 

\subsection{Permutational invariance groups}
As discussed in \citet{niroomand2022capacity}, the magnitude of individual weights must always be viewed with caution due to permutational isomers. For a given neural network of $H$ hidden layers, with $n_l$ nodes in hidden layer $l$, there exist at least $|\mathcal{G}|=\prod_{l=1}^H\left(n_{l} ! \times 2^{n_l}\right)$ sets of weights that are invariant with respect to the model prediction. This effect must be considered when identifying conserved weights; for example, a negative inverse could still be valid and conserved \citep{niroomand2022capacity}. We account for this effect by identifying permutationally invariant sets of weights and only considering a single minimum $m \in \mathcal{G}$ for each $\mathcal{G}$. 

\section{Discussion and conclusions}
Well-established methods from computational chemical physics can be employed to enhance our understanding of machine learning systems. In this work, we have shown how both concepts and associated tools from the study of energy landscapes can be employed for ML-LLs to guide interpretability. We have shown that groups of minima share conserved weights and importantly, that these weights are critical to model performance. Randomly permuting the conserved weights strongly decreases model performance, much more so than permuting any other random set of weights $\mathcal{S}$ of equivalent cardinality $|\mathcal{S}|$. Figure \ref{fig:cc_dg} indicates that all the conserved weights are associated with the particular input node 6. Since the credit card dataset is anonymised and PCA-reduced \citep{dal2015calibrating}, we are unable to say which specific feature it is that helps the model in making a decision, but we can say where it can be found. In Figure \ref{fig:cb_dg}, we know that both input nodes are relevant, which is confirmed by studying the conserved weights for the three given examples. Importantly, different weights are conserved across different examples, highlighting the importance of studying the loss landscape. Studying the applicability of our method to larger and more complex architectures, and perhaps also to different types of machine leaning models, will provide valuable insights, and is an interesting direction for future work.  

\section{Acknowledgements}
DJW gratefully acknowledges an International Chair at the Interdisciplinary Institute for Artificial Intelligence at 3iA Cote d’Azur, supported by the French government, reference ANR-19-P3IA-0002, which has provided interactions that furthered the present research project. MPN acknowledges funding from Downing College, Cambridge.

\bibliographystyle{iclr2023_workshop}
\bibliography{iclr2023_workshop}

\end{document}